\documentclass[10pt,a4paper]{llncs}
\usepackage{cmap} 
\usepackage[british]{babel} 
\usepackage[utf8]{inputenc}
\usepackage{amsmath}
\usepackage[T1]{fontenc}
\usepackage{graphicx, multirow}
\usepackage{algorithm, algpseudocode}
\usepackage{txfonts}
\usepackage{units}
\usepackage{tabularx}
\usepackage[misc]{ifsym}
\usepackage[breaklinks=true,colorlinks,bookmarks=false]{hyperref}
\usepackage{microtype} 
\usepackage{booktabs} 
\usepackage{xspace} 
\usepackage{fancyhdr}

\DeclareMathOperator{\tr}{tr} 
\DeclareMathOperator{\rank}{rank} 
\DeclareMathOperator{\diag}{diag} 
\DeclareMathOperator{\argmin}{argmin} 

\newcommand{\gC}{\ensuremath{C}} 
\newcommand{\gCRIT}{\ensuremath{\mathcal{J}}} 
\newcommand{\gCRITf}[1]{\ensuremath{\gCRIT\left(#1\right)}} 
\newcommand{\gD}{\ensuremath{D}} 
\newcommand{\gDELTA}{\ensuremath{\gB{\Delta}}} 
\newcommand{\gDELTAccH}[2]{\ensuremath{\gH{\delta}\left(#1,#2\right)}} 
\newcommand{\gG}{\ensuremath{\mathcal{G}}} 
\newcommand{\gGn}[1]{\ensuremath{\gB{g}_{#1}}} 
\newcommand{\gGnH}[1]{\ensuremath{\gH{\gB{g}}_{#1}}} 
\newcommand{\gGnc}[2]{\ensuremath{\gGn{#1}^{(#2)}}} 
\newcommand{\gGAMMAjt}[2]{\ensuremath{\gamma_{#1}\left(#2\right)}} 
\newcommand{\gI}{\ensuremath{\gB{I}}} 
\newcommand{\gIc}[1]{\ensuremath{\mathcal{I}_{#1}}} 
\newcommand{\gJ}{\ensuremath{J}} 
\newcommand{\gLAG}[2]{\ensuremath{\mathcal{L}\left(#1,#2\right)}} 
\newcommand{\gLAMBDA}{\ensuremath{\gB{\Lambda}}} 
\newcommand{\gLAMBDAd}[1]{\ensuremath{\lambda_{#1}}} 
\newcommand{\gLAMBDAn}[1]{\ensuremath{\ell_{#1}}} 
\newcommand{\gM}{\ensuremath{\mu}} 
\newcommand{\gMc}[1]{\ensuremath{\gM_{#1}}} 
\newcommand{\gMcH}[1]{\ensuremath{\gH{\gM}_{#1}}} 
\newcommand{\gN}{\ensuremath{N}} 
\newcommand{\gNc}[1]{\ensuremath{\gN_{#1}}} 
\newcommand{\gPHI}{\ensuremath{\gB{\Phi}}} 
\newcommand{\gPHId}[1]{\ensuremath{\gB{f}_{#1}}} 
\newcommand{\gPSI}{\ensuremath{\gB{\Psi}}} 
\newcommand{\gSIGMAc}[1]{\ensuremath{\gB{\Sigma}_{#1}}} 
\newcommand{\gSIGMAcH}[1]{\ensuremath{\gB{\gH{\Sigma}}_{#1}}} 
\newcommand{\gSIGMAb}{\ensuremath{\gSIGMAc{\mathrm{B}}}} 
\newcommand{\gSIGMAw}{\ensuremath{\gSIGMAc{\mathrm{W}}}} 
\newcommand{\gSIGMAt}{\ensuremath{\gSIGMAc{\mathrm{T}}}} 
\newcommand{\gSIGMAtH}{\ensuremath{\gSIGMAcH{\mathrm{T}}}} 
\newcommand{\gT}{\ensuremath{T}} 

\newcommand{\gTHETA}{\ensuremath{\gB{\Theta}}}
\newcommand{\gXI}{\ensuremath{\gB{\Xi}}}
\newcommand{\gUPSILON}{\ensuremath{\gB{\Upsilon}}}

\newcommand{\gCHI}{\ensuremath{\gB{X}}}

\newcommand{\gOMEGA}{\ensuremath{\gB{\Omega}}}

\newcommand{\gB}[1]{\ensuremath{\mathbf{#1}}} 
\newcommand{\gH}[1]{\ensuremath{\widehat{#1}}} 
\newcommand{\gL}[1]{\ensuremath{{#1}_L}} 
\newcommand{\gE}[1]{\ensuremath{{#1}_E}} 

\let\exper\textbf

\title{Walker-Independent~Features for~Gait~Recognition from~Motion~Capture~Data}

\author{Michal Balazia (\href{https://orcid.org/0000-0001-7153-9984}{0000-0001-7153-9984}) \and Petr Sojka (\href{https://orcid.org/0000-0002-5768-4007}{0000-0002-5768-4007})}
\institute{Faculty of Informatics, Masaryk University,
Botanick\'a 68a, 602\,00 Brno, Czech Republic\\
\email{xbalazia@mail.muni.cz} and \email{sojka@fi.muni.cz}
}

\begin{document}

\maketitle

\pagestyle{plain}
\thispagestyle{fancy}
\fancyhead[C]{Joint International Workshops on Structural and Syntactic Pattern Recognition\\and Statistical Techniques in Pattern Recognition 2016, preprint}
\headheight23pt

\begin{abstract}
MoCap-based human identification, as a pattern recognition discipline, can be optimized using a machine learning approach. Yet in some applications such as video surveillance new identities can appear on the fly and labeled data for all encountered people may not always be available. This work introduces the concept of learning walker-independent gait features directly from raw joint coordinates by a modification of the Fisher’s Linear Discriminant Analysis with Maximum Margin Criterion. Our new approach shows not only that these features can discriminate different people than who they are learned on, but also that the number of learning identities can be much smaller than the number of walkers encountered in the real operation.
\end{abstract}

\section{Introduction}
\label{intro}

\begin{figure}[h]
\vspace{-6pt}%
\centering
\includegraphics[height=4cm]{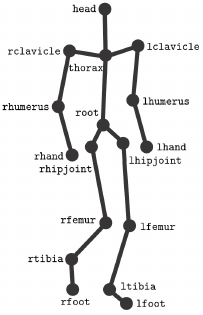}\qquad
\includegraphics[height=4cm]{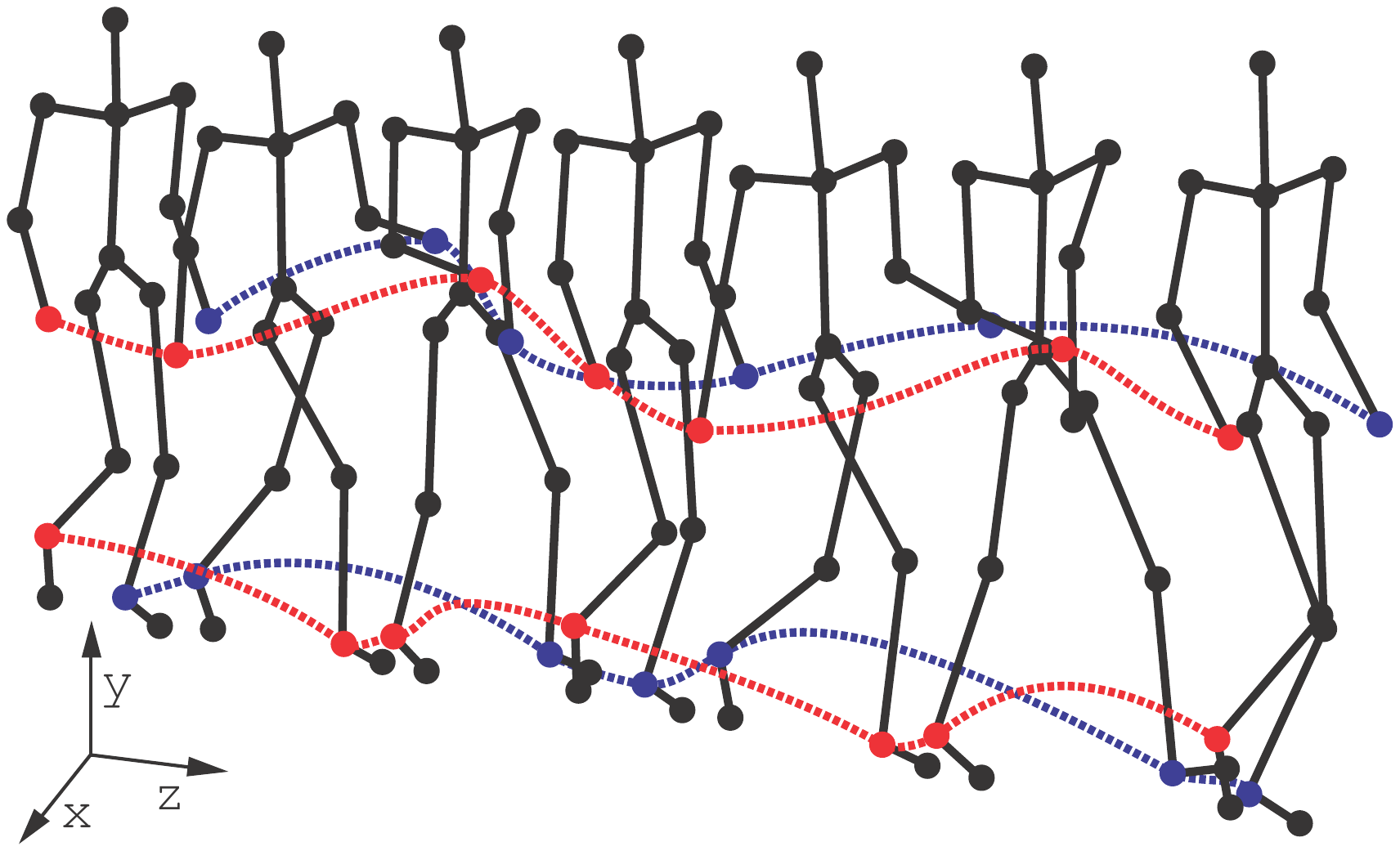}
\vspace{-6pt}%
\caption{Motion capture data. Skeleton is represented by a stick figure of 31~joints (only 17 are drawn here). Seven selected video frames of a walk sequence contain 3D coordinates of each joint in time. The red and blue lines track trajectories of hands and feet.~\cite{VBZ16}}
\label{f1}
\end{figure}

Recent rapid improvement in motion capture (MoCap) sensor accuracy brought affordable technology that can identify walking people. MoCap technology provides video clips of walking individuals containing structural motion data. The format keeps an overall structure of the human body and holds estimated 3D positions of major anatomical landmarks as the person moves. MoCap data can be collected online by a system of multiple cameras (Vicon) or a depth camera (Microsoft Kinect). To visualize motion capture data (see Figure~\ref{f1}), a simplified stick figure representing the human skeleton (graph of joints connected by bones) can be recovered from body point spatial coordinates. 

Recognizing a person by walk involves capturing and normalizing their walk sample, extracting gait features to compose a template, and finally querying a central database for a set of similar templates to report the most likely identity. This work focuses on extracting robust and discriminative gait features from raw MoCap data.

Many geometric gait features have been introduced over the past few years. They are typically combinations of static body parameters (bone lengths, person's height)~\cite{KKMJ14} with dynamic gait features such as step length, walk speed, joint angles and inter-joint distances~\cite{AA15,APG15,KKMJ14,RCA15}, along with various statistics (mean, standard deviation or maximum) of their signals~\cite{AWLSWZ16}. Clearly, these features are schematic and human-interpretable, which is convenient for visualizations and for intuitive understanding, but unnecessary for automatic gait recognition. Instead, this application prefers learning features that maximally separate the identity classes and are not limited by such dispensable factors.

Methods for 2D~gait recognition extensively use machine learning models for extracting gait features, such as principal component analysis and multi-scale shape analysis~\cite{CT15}, genetic algorithms and kernel principal component analysis~\cite{TBLH15}, radial basis function neural networks~\cite{ZW14}, or convolutional neural networks~\cite{CMGP16}. All of those and many other models are reasonable to be utilized also in 3D~gait recognition.

In the video surveillance environment data need to be acquired without walker's consent and new identities can appear on the fly. Here and also in other applications where labels for all encountered people may not always be available, we value features that have a high power in distinguishing all people and not exclusively who they were learned on. We call these walker-independent features. The main idea is to statistically learn what aspects of walk people generally differ in and extract those as gait features. The features are learned in a supervised manner, as described in the following section.

\section{Learning Gait Features}
\label{meth}

In statistical pattern recognition, reducing space dimensionality is a common technique to overcome class estimation problems. Classes are discriminated by projecting high-dimensional input data onto low-dimensional sub-spaces by linear transformations with the goal of maximizing the class separability. We are interested in finding an optimal feature space where a gait template is close to those of the same walker and far from those of different walkers.

Let the model of a human body have $\gJ$ joints and all samples be linearly normalized to their average length~$\gT$. Labeled learning data in the measurement space $\gL{\gG}$ are in the form $\left\{\left(\gGn{n},\gLAMBDAn{n}\right)\right\}_{n=1}^{\gL{\gN}}$ where
\begin{equation}
\gGn{n}=\left[[\gGAMMAjt{1}{1}\,\cdots\,\gGAMMAjt{\gJ}{1}]^\top\,\cdots\,[\gGAMMAjt{1}{\gT}\,\cdots\,\gGAMMAjt{\gJ}{\gT}]^\top\right]^\top
\end{equation}
is a gait sample (one gait cycle) in which $\gGAMMAjt{j}{t}\in\mathbb{R}^3$ are 3D spatial coordinates of a joint $j\in\left\{1,\ldots,\gJ\right\}$ at time $t\in\left\{1,\ldots,\gT\right\}$ normalized with respect to the person's position and walk direction. See that $\gL{\gG}$ has dimensionality $\gD=3\gJ\gT$. Each learning sample falls strictly into one of the learning identity classes $\left\{\gIc{c}\right\}_{c=1}^{\gC}$ determined by $\gLAMBDAn{n}$. A class $\gIc{c}\subseteq\gL{\gG}$ has $\gNc{c}$ samples. The classes are complete and mutually exclusive. We say that learning samples $\left(\gGn{n},\gLAMBDAn{n}\right)$ and $\left(\gGn{n'},\gLAMBDAn{n'}\right)$ share a common walker if and only if they belong to the same class, i.e., $\left(\gGn{n},\gLAMBDAn{n}\right),\left(\gGn{n'},\gLAMBDAn{n'}\right)\in\gIc{c}\Leftrightarrow\gLAMBDAn{n}=\gLAMBDAn{n'}$.

We measure class separability of a given feature space by a representation of the Maximum Margin Criterion (MMC)~\cite{KKS04,LJZ06} used by the Vapnik's Support Vector Machines (SVM)~\cite{V95}
\begin{equation}
\gCRIT=\frac{1}{2}\sum_{c,c'=1}^{\gL{\gC}}\left(\left(\gMc{c}-\gMc{c'}\right)^\top\left(\gMc{c}-\gMc{c'}\right)-\tr\left(\gSIGMAc{c}+\gSIGMAc{c'}\right)\right)
\end{equation}
which is actually a summation of $\frac{1}{2}\gL{\gC}(\gL{\gC}-1)$ between-class margins. The margin is defined as the Euclidean distance of class means minus both individual variances (traces of scatter matrices $\gSIGMAc{c}=\frac{1}{\gNc{c}}\sum_{n=1}^{\gNc{c}}\left(\gGnc{n}{c}-\gMc{c}\right)\left(\gGnc{n}{c}-\gMc{c}\right)^\top$ and similarly for $\gSIGMAc{c'}$). For the whole labeled data, we denote the between- and within-class and total scatter matrices
\begin{equation}
\begin{split}
\gSIGMAb & =\sum_{c=1}^{\gL{\gC}}\left(\gMc{c}-\gM\right)\left(\gMc{c}-\gM\right)^\top\\
\gSIGMAw & =\sum_{c=1}^{\gL{\gC}}\frac{1}{\gNc{c}}\sum_{n=1}^{\gNc{c}}\left(\gGnc{n}{c}-\gMc{c}\right)\left(\gGnc{n}{c}-\gMc{c}\right)^\top\\
\gSIGMAt & =\sum_{c=1}^{\gL{\gC}}\frac{1}{\gNc{c}}\sum_{n=1}^{\gNc{c}}\left(\gGnc{n}{c}-\gM\right)\left(\gGnc{n}{c}-\gM\right)^\top=\gSIGMAb+\gSIGMAw
\end{split}
\end{equation}
where $\gGnc{n}{c}$ denotes the $n$-th sample in class $\gIc{c}$ and $\gMc{c}$ and $\gM$ are sample means for class $\gIc{c}$ and the whole data set, respectively, that is, $\gMc{c}=\frac{1}{\gNc{c}}\sum_{n=1}^{\gNc{c}}\gGnc{n}{c}$ and $\gM=\frac{1}{\gL{\gN}}\sum_{n=1}^{\gL{\gN}}\gGn{n}$. Now we obtain
\begin{equation}
\begin{split}
\gCRIT & =\frac{1}{2}\sum_{c,c'=1}^{\gL{\gC}}\left(\gMc{c}-\gMc{c'}\right)^\top\left(\gMc{c}-\gMc{c'}\right)-\frac{1}{2}\sum_{c,c'=1}^{\gL{\gC}}\tr\left(\gSIGMAc{c}+\gSIGMAc{c'}\right)\\
& =\frac{1}{2}\sum_{c,c'=1}^{\gL{\gC}}\left(\gMc{c}-\gM+\gM-\gMc{c'}\right)^\top\left(\gMc{c}-\gM+\gM-\gMc{c'}\right)-\sum_{c=1}^{\gL{\gC}}\tr\left(\gSIGMAc{c}\right)\\
& =\tr\left(\sum_{c=1}^{\gL{\gC}}\left(\gMc{c}-\gM\right)\left(\gMc{c}-\gM\right)^\top\right)-\tr\left(\sum_{c=1}^{\gL{\gC}}\gSIGMAc{c}\right)\\
& =\tr\left(\gSIGMAb\right)-\tr\left(\gSIGMAw\right)=\tr\left(\gSIGMAb-\gSIGMAw\right).
\end{split}
\end{equation}
Since $\tr\left(\gSIGMAb\right)$ measures the overall variance of the class mean vectors, a large one implies that the class mean vectors scatter in a large space. On the other hand, a small $\tr\left(\gSIGMAw\right)$ implies that classes have a small spread. Thus, a large $\gCRIT$ indicates that samples are close to each other if they share a common walker but are far from each other if they are performed by different walkers. Extracting features, that is, transforming the input data in the measurement space into a feature space of higher $\gCRIT$, can be used to link new observations of walkers more successfully.
\pagebreak

Feature extraction is given by a linear transformation (feature) matrix $\gPHI\in\mathbb{R}^{\gD\times\gH{\gD}}$ from a $\gD$-dimensional measurement space $\gG=\left\{\gGn{n}\right\}_{n=1}^{\gN}$ of not necessarily labeled gait samples to a $\gH{\gD}$-dimensional feature space $\gH{\gG}=\left\{\gGnH{n}\right\}_{n=1}^{\gN}$ of gait templates where $\gH{\gD}<\gD$ and each gait sample $\gGn{n}$ is transformed into a gait template $\gGnH{n}=\gPHI^\top\gGn{n}$. The objective is to learn a transform $\gPHI$ that maximizes MMC in the feature space
\begin{equation}
\gCRITf{\gPHI}=\tr\left(\gPHI^\top\left(\gSIGMAb-\gSIGMAw\right)\gPHI\right).
\label{e2}
\end{equation}
Once the transformation is found, all measured samples are transformed into templates (in the feature space) along with the class means and covariances. The templates are compared by the Mahalanobis distance function
\begin{equation}
\gDELTAccH{\gGnH{n}}{\gGnH{n'}}=\sqrt{\left(\gGnH{n}-\gGnH{n'}\right)^\top\gSIGMAtH^{-1}\left(\gGnH{n}-\gGnH{n'}\right)}.
\label{e3}
\end{equation}

We show that solution to the optimization problem in Equation~\eqref{e2} can be obtained by eigendecomposition of the matrix $\gSIGMAb-\gSIGMAw$. An important property to notice about the objective $\gCRITf{\gPHI}$ is that it is invariant w.r.t.\@ rescalings $\gPHI\rightarrow\alpha\gPHI$. Hence, we can always choose $\gPHI=\gPHId{1}\|\cdots\|\gPHId{\gH{\gD}}$ such that $\gPHId{\gH{d}}^\top\gPHId{\gH{d}}=1$, since it is a scalar itself. For this reason we can reduce the problem of maximizing $\gCRITf{\gPHI}$ into the constrained optimization problem
\begin{equation}
\begin{split}
\max & \quad\sum_{\gH{d}=1}^{\gH{\gD}}\gPHId{\gH{d}}^\top\left(\gSIGMAb-\gSIGMAw\right)\gPHId{\gH{d}}\\
\mathrm{subject\,to} & \quad\gPHId{\gH{d}}^\top\gPHId{\gH{d}}-1=0\qquad\forall\gH{d}=1,\ldots,\gH{\gD}.
\end{split}
\end{equation}
To solve the above optimization problem, let us consider the Lagrangian
\begin{equation}
\gLAG{\gPHId{\gH{d}}}{\gLAMBDAd{\gH{d}}}=\sum_{\gH{d}=1}^{\gH{\gD}}\gPHId{\gH{d}}^\top\left(\gSIGMAb-\gSIGMAw\right)\gPHId{\gH{d}}-\gLAMBDAd{\gH{d}}\left(\gPHId{\gH{d}}^\top\gPHId{\gH{d}}-1\right)
\end{equation}
with multipliers $\gLAMBDAd{\gH{d}}$. To find the maximum, we derive it with respect to $\gPHId{\gH{d}}$ and equate to zero
\begin{equation}
\frac{\partial\gLAG{\gPHId{\gH{d}}}{\gLAMBDAd{\gH{d}}}}{\partial\gPHId{\gH{d}}}=\left(\left(\gSIGMAb-\gSIGMAw\right)-\gLAMBDAd{\gH{d}}\gI\right)\gPHId{\gH{d}}=0
\end{equation}
which leads to
\begin{equation}
\left(\gSIGMAb-\gSIGMAw\right)\gPHId{\gH{d}}=\gLAMBDAd{\gH{d}}\gPHId{\gH{d}}
\end{equation}
where $\gLAMBDAd{\gH{d}}$ are the eigenvalues of $\gSIGMAb-\gSIGMAw$ and $\gPHId{\gH{d}}$ are the corresponding eigenvectors. Putting it all together,
\begin{equation}
\left(\gSIGMAb-\gSIGMAw\right)\gPHI=\gLAMBDA\gPHI
\end{equation}
where $\gLAMBDA=\diag\left(\gLAMBDAd{1},\ldots,\gLAMBDAd{\gH{\gD}}\right)$ is the eigenvalue matrix. Therefore,
\begin{equation}
\gCRITf{\gPHI}=\tr\left(\gPHI^\top\left(\gSIGMAb-\gSIGMAw\right)\gPHI\right)=\tr\left(\gPHI^\top\gLAMBDA\gPHI\right)=\tr\left(\gLAMBDA\right)
\end{equation}
is maximized when $\gLAMBDA$ has $\gH{\gD}$ largest eigenvalues and $\gPHI$ contains the corresponding leading eigenvectors.

In the following we discuss how to calculate the eigenvectors of $\gSIGMAb-\gSIGMAw$ and to determine an optimal dimensionality $\gH\gD$ of the feature space. Rewrite $\gSIGMAb-\gSIGMAw=2\gSIGMAb-\gSIGMAt$. Note that the null space of $\gSIGMAt$ is a subspace of that of $\gSIGMAb$ since the null space of $\gSIGMAt$ is the common null space of $\gSIGMAb$ and $\gSIGMAw$. Thus, we can simultaneously diagonalize $\gSIGMAb$ and $\gSIGMAt$ to some $\gDELTA$ and $\gI$
\begin{equation}
\begin{split}
\gPSI^\top\gSIGMAb\gPSI & =\gDELTA\\
\gPSI^\top\gSIGMAt\gPSI & =\gI
\end{split}
\end{equation}
with the $\gD\times\rank\left(\gSIGMAt\right)$ eigenvector matrix
\begin{equation}
\gPSI=\gOMEGA\gTHETA^{-\frac{1}{2}}\gXI
\end{equation}
where $\gOMEGA$ and $\gTHETA$ are the eigenvector and eigenvalue matrices of $\gSIGMAt$, respectively, and $\gXI$ is the eigenvector matrix of $\gTHETA^{-1/2}\gOMEGA^\top\gSIGMAb\gOMEGA\gTHETA^{-1/2}$. To calculate $\gPSI$, we use a fast two-step algorithm in virtue of Singular Value Decomposition (SVD). SVD expresses a real $r \times s$ matrix $\gB{A}$ as a product $\gB{A}=\gB{U}\gB{D}\gB{V}^\top$ where $\gB{D}$ is a diagonal matrix with decreasing non-negative entries, and $\gB{U}$ and $\gB{V}$ are $r\times\min\left\{r,s\right\}$ and $s\times\min\left\{r,s\right\}$ eigenvector matrices of $\gB{A}\gB{A}^\top$ and $\gB{A}^\top\gB{A}$, respectively, and the non-vanishing entries of $\gB{D}$ are square roots of the non-zero corresponding eigenvalues of both $\gB{A}\gB{A}^\top$ and $\gB{A}^\top\gB{A}$. See that $\gSIGMAt$ and $\gSIGMAb$ can be expressed in the forms
\vspace{-4pt}%
\begin{equation}
\begin{split}
\gSIGMAt=&\enskip\gCHI\gCHI^\top\enskip\mathrm{where}\enskip\gCHI=\frac{1}{\sqrt{\gL{\gN}}}\left[\left(\gGn{1}-\gM\right)\cdots\left(\gGn{\gL{\gN}}-\gM\right)\right]\enskip\text{and}\\
\gSIGMAb=&\enskip\gUPSILON\gUPSILON^\top\enskip\text{where}\enskip\gUPSILON=\left[\left(\gMc{1}-\gM\right)\cdots\left(\gMc{\gL{\gC}}-\gM\right)\right],
\end{split}
\end{equation}
respectively. Hence, we can obtain the eigenvectors $\gOMEGA$ and the corresponding eigenvalues $\gTHETA$ of $\gSIGMAt$ through the SVD of $\gCHI$ and analogically $\gXI$ of $\gTHETA^{-1/2}\gOMEGA^\top\gSIGMAb\gOMEGA\gTHETA^{-1/2}$ through the SVD of $\gTHETA^{-1/2}\gOMEGA^\top\gUPSILON$. The columns of $\gPSI$ are clearly the eigenvectors of $2\gSIGMAb-\gSIGMAt$ with the corresponding eigenvalues $2\gDELTA-\gI$. Therefore, to constitute the transform $\gPHI$ by maximizing the MMC, we should choose the eigenvectors in $\gPSI$ that correspond to the eigenvalues of at least $\frac{1}{2}$ in $\gDELTA$. Note that $\gDELTA$ contains at most $\rank\left(\gSIGMAb\right)=\gL{\gC}-1$ positive eigenvalues, which gives an upper bound on the feature space dimensionality~$\gH\gD$. Algorithm~\ref{a1}~\cite{BS16a} provided below is an efficient way of learning the transform $\gPHI$ for MMC on given labeled learning data~$\gL{\gG}$.

\vspace{-20pt}%
\begin{algorithm}[b]
\caption{LearnTransformationMatrixMMC$\left(\gL{\gG}\right)$}
\label{a1}
\begin{algorithmic}[1]
  \State split $\gL{\gG}=\left\{\left(\gGn{n},\gLAMBDAn{n}\right)\right\}_{n=1}^{\gL{\gN}}$ into classes $\left\{\gIc{c}\right\}_{c=1}^{\gL{\gC}}$ of $\gNc{c}=\left|\gIc{c}\right|$ samples
  \State compute overall mean $\gM=\frac{1}{\gL{\gN}}\sum_{n=1}^{\gL{\gN}}\gGn{n}$ and individual class means $\gMc{c}=\frac{1}{\gNc{c}}\sum_{n=1}^{\gNc{c}}\gGnc{n}{c}$
  \State compute $\gSIGMAb=\sum_{c=1}^{\gL{\gC}}\left(\gMc{c}-\gM\right)\left(\gMc{c}-\gM\right)^\top$
  \State compute $\gCHI=\frac{1}{\sqrt{\gL{\gN}}}\left[\left(\gGn{1}-\gM\right)\cdots\left(\gGn{\gL{\gN}}-\gM\right)\right]$
  \State compute $\gUPSILON=\left[\left(\gMc{1}-\gM\right)\cdots\left(\gMc{\gL{\gC}}-\gM\right)\right]$
  \State compute eigenvectors $\gOMEGA$ and corresponding eigenvalues $\gTHETA$ of $\gSIGMAt$ through SVD of $\gCHI$
  \State compute eigenvectors $\gXI$ of $\gTHETA^{\nicefrac{-1}{2}}\gOMEGA^\top\gSIGMAb\gOMEGA\gTHETA^{\nicefrac{-1}{2}}$ through SVD of $\gTHETA^{\nicefrac{-1}{2}}\gOMEGA^\top\gUPSILON$
  \State compute eigenvectors $\gPSI=\gOMEGA\gTHETA^{\nicefrac{-1}{2}}\gXI$
  \State compute eigenvalues $\gDELTA=\gPSI^\top\gSIGMAb\gPSI$
  \State return transform $\gPHI$ as eigenvectors in $\gPSI$ that correspond to the eigenvalues of at least $\nicefrac{1}{2}$ in $\gDELTA$
\end{algorithmic}
\end{algorithm}

\section{Experiments and Results}
\label{exp}

\subsection{Database}
\label{exp-db}

For the evaluation purposes we have extracted a large number of samples from the general MoCap database from CMU~\cite{CMU03} as a well-known and recognized database of structural human motion data. It contains numerous motion sequences, including a considerable number of gait sequences. Motions are recorded with an optical marker-based Vicon system. People wear a black jumpsuit and have 41~markers taped on. The tracking space of \unit[30]{m$^2$}, surrounded by 12~cameras of sampling rate of \unit[120]{Hz} in the height from 2 to 4~meters above ground, creates a video surveillance environment. Motion videos are triangulated to get highly accurate 3D data in the form of relative body point coordinates (with respect to the root joint) in each video frame and stored in the standard ASF/AMC data format. Each registered participant is assigned with their respective skeleton described in an ASF file. Motions in the AMC files store bone rotational data, which is interpreted as instructions about how the associated skeleton deforms over time.

These MoCap data, however, contain skeleton parameters pre-calibrated by the CMU staff. Skeletons are unique for each walker and even a trivial skeleton check could result in 100\% recognition. In order to use the collected data in a fairly manner, a prototypical skeleton is constructed and used to represent bodies of all subjects, shrouding the unique skeleton parameters of individual walkers. Assuming that all walking subjects are physically identical disables the skeleton check as a potentially unfair classifier. Moreover, this is a skeleton-robust solution as all bone rotational data are linked with a~fixed skeleton. To obtain realistic parameters, it is calculated as the mean of all skeletons in the provided ASF files.

We calculate 3D joint coordinates using bone rotational data and the prototypical skeleton. One cannot directly use raw values of joint coordinates, as they refer to absolute positions in the tracking space, and not all potential methods are invariant to person's position or walk direction. To ensure such invariance, the center of the coordinate system is moved to the position of root joint $\gGAMMAjt{\mathrm{root}}{t}=[0,0,0]^\top$ for each time~$t$ and axes are adjusted to the walker's perspective: the X~axis is from right (negative) to left (positive), the Y~axis is from down (negative) to up (positive), and the Z~axis is from back (negative) to front (positive). In the AMC file structure notation it is achieved by zeroing the root translation and rotation (\texttt{root 0 0 0 0 0 0}) in all frames of all motion sequences.

Since the general motion database contains all motion types, we extracted a number of sub-motions that represent gait cycles. First, an exemplary gait cycle was identified, and clean gait cycles were then filtered out using the DTW distance over bone rotations. The similarity threshold was set high enough so that even the least similar sub-motion still semantically represents a gait cycle. Finally, subjects that contributed with less than 10~samples were excluded. The final database has 54~walking subjects that performed 3{,}843~samples in total, which makes an average of about 71~samples per subject.

\subsection{Evaluation Setups and Metrics}
\label{exp-su}
\begin{figure}[h]
\vspace{-5pt}%
\centering
\includegraphics[width=0.8\textwidth]{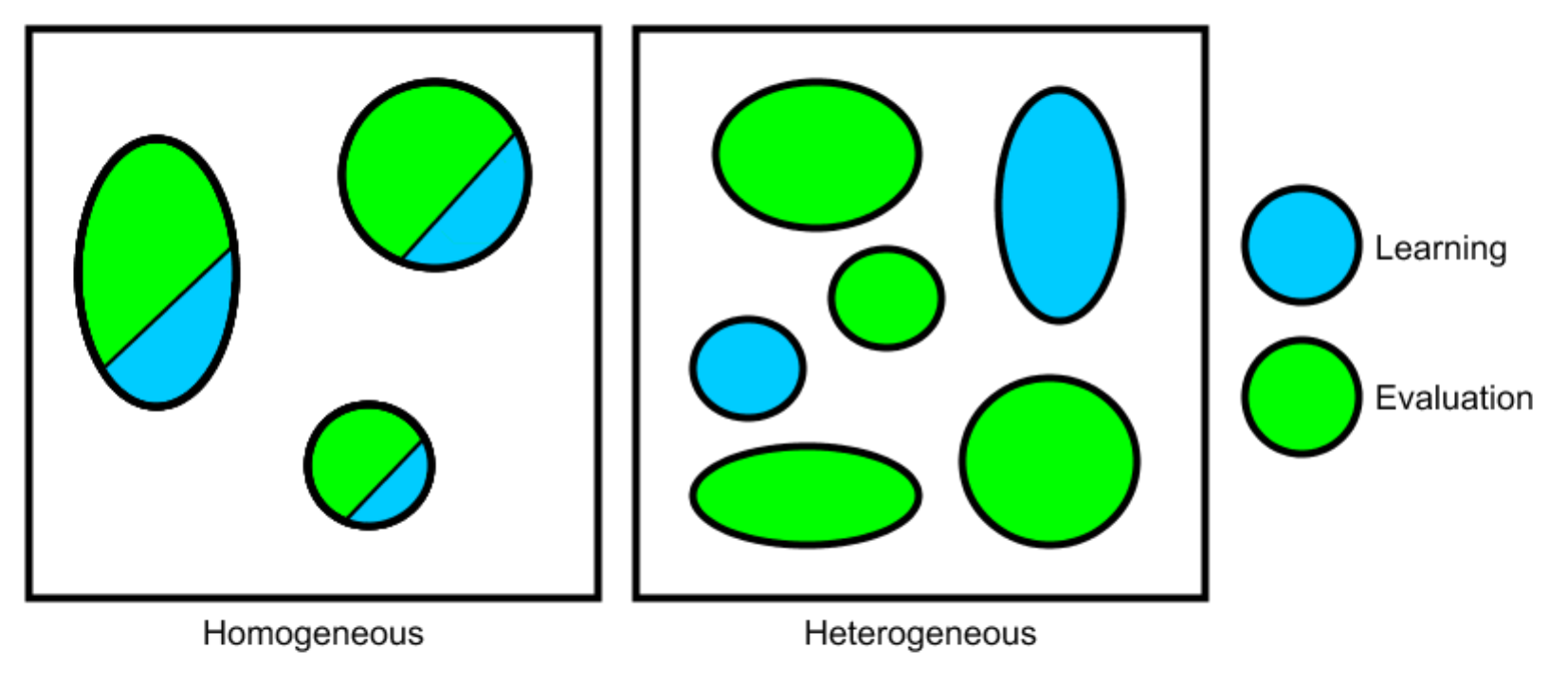}
\vspace{-5pt}%
\caption{Abstraction of data separation for homogeneous setup of $\gL{\gC}=\gE{\gC}=3$ learning-and-evaluation classes (left) and for heterogeneous setup of~$\gL{\gC}=2$ learning classes and $\gE{\gC}=4$ evaluation classes (right). Black square represents a database and ellipses are identity classes.}
\label{f2}
\vspace{-2pt}%
\end{figure}

Learning data $\gL{\gG}=\left\{\left(\gGn{n},\gLAMBDAn{n}\right)\right\}_{n=1}^{\gL{\gN}}$ of $\gL{\gC}$ identities and evaluation data $\gE{\gG}=\left\{\left(\gGn{n},\gLAMBDAn{n}\right)\right\}_{n=1}^{\gE{\gN}}$ of $\gE{\gC}$ identity classes have to be disjunct at all times. Evaluation is performed exclusively on the evaluation part, taking no observations of the learning part into account. In the following we introduce two setups of data separation: homogeneous and heterogeneous. The homogeneous setup learns the transformation matrix on $\nicefrac{1}{3}$ samples of $\gL{\gC}$ identities and is evaluated on templates derived from other $\nicefrac{2}{3}$ samples of the same $\gE{\gC}=\gL{\gC}$ identities. The heterogeneous setup learns the transform on all samples in $\gL{\gC}$ identities and is evaluated on all templates derived from other $\gE{\gC}$ identities. For better clarification we refer to Figure~\ref{f2}. Note that unlike in homogeneous setup, in heterogeneous setup there is no walker identity ever used for both learning and evaluation at the same time.

Homogeneous setup is parametrized by a single number $\gL{\gC}=\gE{\gC}$ of learning-and-evaluation identity classes, whereas the heterogeneous setup has the form $\left(\gL{\gC},\gE{\gC}\right)$ specifying how many learning and how many evaluation identity classes are randomly selected from the database. Evaluation of each setup is repeated 3~times, selecting new random $\gL{\gC}$ and $\gE{\gC}$ identity classes each time and reporting the average result. Please note that in the heterogeneous setup the learning identities are disjunct from the evaluation identities, that is, there is no single identity used for both learning and evaluation.

Correct Classification Rate (CCR) is a standard qualitative measure; however, if a~method has a low CCR, we cannot directly say if the system is failing because of bad features or a bad classifier. Providing an evaluation in terms of class separability of the feature space gives an estimate on the recognition potential of the extracted features and do not reflect eventual combination with an unsuitable classifier. Quality of features extraction algorithms is reflected in the Davies-Bouldin Index (DBI)
\begin{equation}
\mathrm{DBI}=\frac{1}{\gE{\gC}}\sum_{c=1}^{\gE{\gC}}\max\limits_{1 \leq c'\leq\gE{\gC},\,c' \neq c}\frac{\sigma_c+\sigma_{c'}}{\gDELTAccH{\gMcH{c}}{\gMcH{c'}}}
\end{equation}
where $\sigma_c=\frac{1}{\gNc{c}}\sum_{n=1}^{\gNc{c}}\gDELTAccH{\gGnH{n}}{\gMcH{c}}$ is the average distance of all templates in identity class $\gIc{c}$ to its centroid, and similarly for $\sigma_{c'}$. Templates of low intra-class distances and of high inter-class distances have a low DBI. DBI is measured on the full evaluation part, whereas CCR is estimated with 10-fold cross-validation taking one dis-labeled fold as a~testing set and other nine as gallery. Test templates are classified by the winner-takes-all strategy, in which a test template $\gGnH{}^{\mathrm{test}}$ gets assigned with the label $\gLAMBDAn{\argmin_i\gDELTAccH{\gGnH{}^{\mathrm{test}}}{\gGnH{i}^{\mathrm{gallery}}}}$ of the gallery's closest identity class.

Based on Section~\ref{exp-db}, our database has 54~identity classes in total. We performed the series of experiments \exper{A}, \exper{B}, \exper{C}, \exper{D} below. The experiments \exper{A} and \exper{B} are to compare the homogeneous and heterogeneous setup, whereas \exper{C} and \exper{D} examine how performance of the system in the heterogeneous setup improves with increasing number of learning identities. The results are illustrated in Figure~\ref{f3} and in Figure~\ref{f4} in the next section.
\begin{description}
\item[\exper{A}] homogeneous setup with $\gL{\gC}=\gE{\gC}\in\left\{2,\ldots,27\right\}$;
\item[\exper{B}] heterogeneous setup with $\gL{\gC}=\gE{\gC}\in\left\{2,\ldots,27\right\}$;
\item[\exper{C}] heterogeneous setup with $\gL{\gC}\in\left\{2,\ldots,27\right\}$ and $\gE{\gC}=27$;
\item[\exper{D}] heterogeneous setup with $\gL{\gC}\in\left\{2,\ldots,52\right\}$ and $\gE{\gC}=54-\gL{\gC}$.
\end{description}

\subsection{Results}
\label{exp-r}

\begin{figure}[h]
\includegraphics[height=3.5cm]{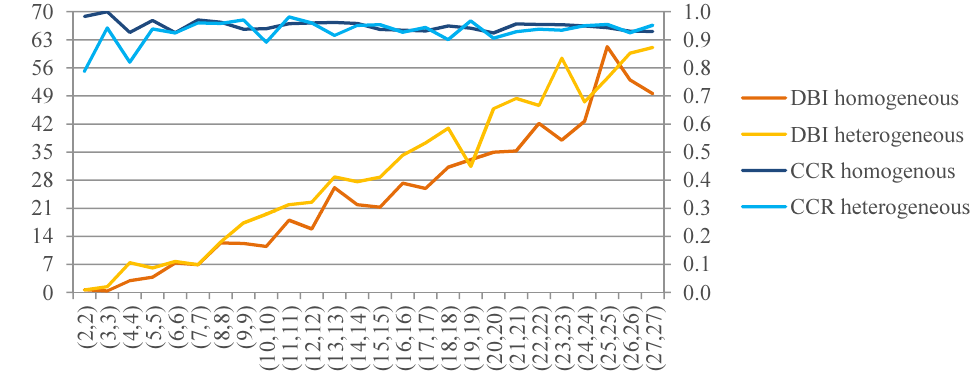}\\
\includegraphics[height=3.5cm]{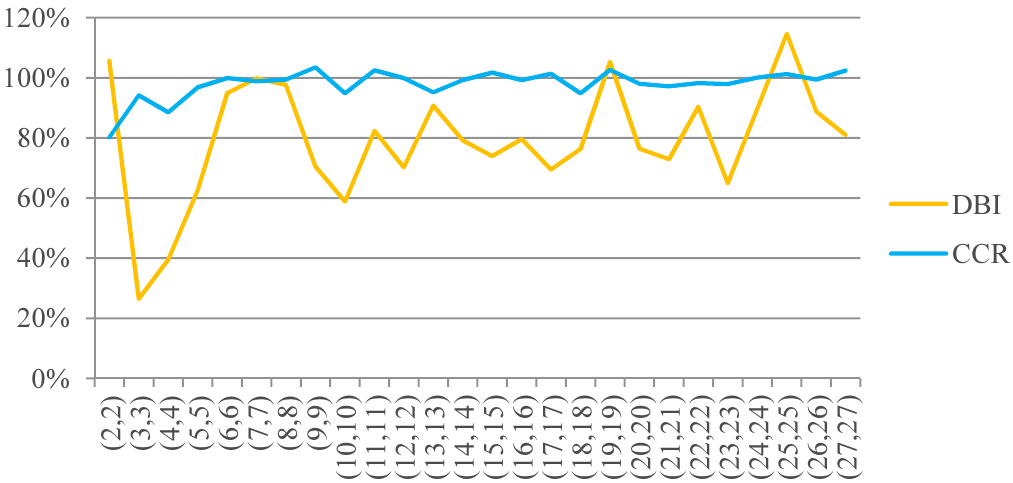}
\caption{DBI (left vertical axis) and CCR (right vertical axis) for experiments \exper{A} of homogeneous setup and \exper{B} of heterogeneous setup (top) with $\left(\gL{\gC},\gE{\gC}\right)$ configurations (horizontal axes) and their percentages (bottom).}
\label{f3}
\end{figure}

Experiments \exper{A} and \exper{B} compare homogeneous and heterogeneous setups by measuring the drop in performance on an identical number of learning and evaluation identities ($\gL{\gC}=\gE{\gC}$). Top plot in Figure~\ref{f3} shows the measured values of DBI and CCR metrics in both alternatives, which not only appear comparable but also in some configurations the heterogeneous setup has an even higher CCR. Bottom plot expresses heterogeneous setup as a percentage of the homogeneous setup in each of the particular metrics. Here we see that with raising number of identities the heterogeneous setup approaches 100\% of the fully homogeneous alternative.

\begin{figure}[h]
\vspace{-1pt}%
\includegraphics[height=3.5cm]{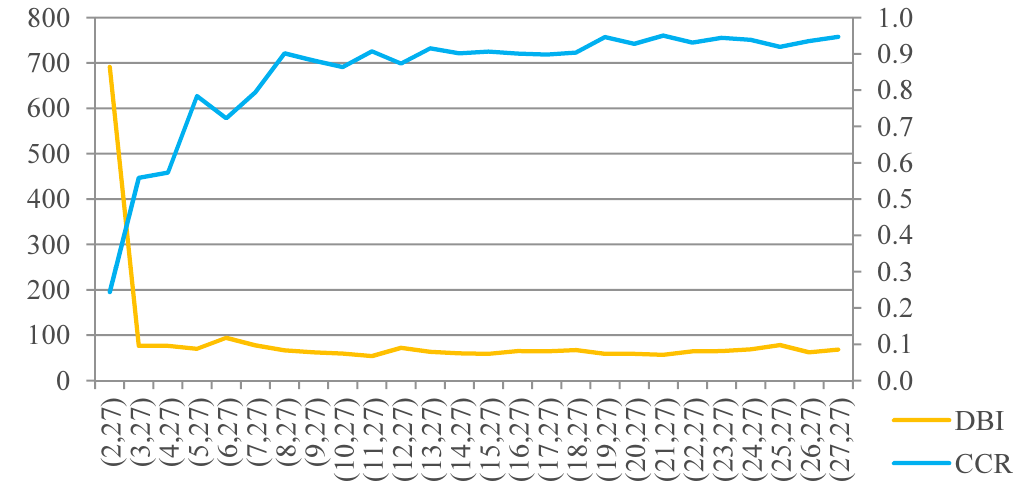}\\
\includegraphics[height=3.5cm]{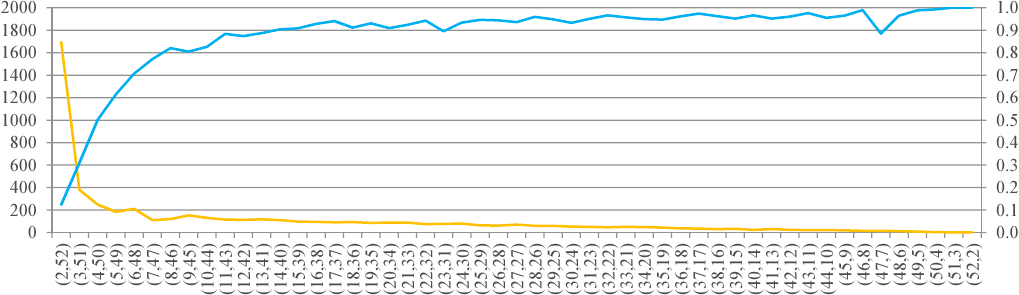}
\caption{DBI (left vertical axes) and CCR (right vertical axes) for experiments \exper{C} (top) and \exper{D} (bottom) on heterogeneous setup with $\left(\gL{\gC},\gE{\gC}\right)$ configurations (horizontal axes).}
\label{f4}
\end{figure}

Experiments \exper{C} and \exper{D} investigate on the impact of the number of learning identities in the heterogeneous setup. Observing from the Figure~\ref{f4}, the performance grows quickly on the first configurations with very few learning identities, which we can interpret as an analogy to the Pareto (80--20) principle. Specifically, the results of experiment \exper{C} say that 8~learning identities achieve almost the same performance (66.78~DBI and 0.902~CCR) to as if learned on 27~identities (68.32~DBI and 0.947~CCR). The outcome of experiment~\exper{D} indicates a similar growth of performance and we see that yet 14~identities can be enough to learn the transformation matrix to distinguish 40 completely different people (0.904~CCR).

The proposed method and seven other state-of-the-art methods~\cite{AAS14,AA15,BRRV12,DMG14,KKMJ14,PKWL12,SCB13} have been subjected to extensive simulations on homogeneous setup in our recent research paper~\cite{BS16a}. A variety of class-separability coefficients and classification metrics allows insights from different statistical perspectives. Results indicate that the proposed method is a leading concept for rank-based classifier systems: lowest Davies-Bouldin Index, highest Dunn Index, highest (and exclusively positive) Silhouette Coefficient, second highest Fisher's Discriminant Ratio and, combined with rank-based classifier, the best Cumulative Match Characteristic, False Accept Rate and False Reject Rate trade-off, Receiver Operating Characteristic (ROC) and recall-precision trade-off scores along with Correct Classification Rate, Equal Error Rate, Area Under ROC Curve and Mean Average Precision. We interpret the high scores as a sign of robustness. Apart from performance merits, the MMC method is also efficient: low-dimensional templates ($\gH{\gD}\leq\gL{\gC}-1=\gE{\gC}-1=53$) and Mahalanobis distance ensure fast distance computations and thus contribute to high scalability.

\section{Conclusions}
\label{con}

Despite many advanced optimization techniques used in statistical pattern recognition, a~common practice of state-of-the-art MoCap-based human identification is still to design geometric gait features by hand. As the first contribution of this paper, the proposed method does not involve any ad-hoc features; on the contrary, they are computed from a~much larger space beyond the limits of human interpretability. The features are learned directly from raw joint coordinates by a~modification of the Fisher's LDA with MMC so that the identities are maximally separated. We believe that MMC is a suitable criterion for optimizing gait features; however, our future work will continue with research on further potential optimality criterions and machine learning approaches. Furthermore, we are in the process of developing an evaluation framework with implementation details and source codes of all related methods, data extraction drive from the general CMU MoCap database and the evaluation mechanism to support reproducible research.

Second contribution lies in showing the possibility of building a representation on a problem and using it on another (related) problem.  Simulations on the CMU MoCap database show that our approach is able to build robust feature spaces without pre-registering and labeling all potential walkers. In fact, we can take different people (experiments \exper{A} and \exper{B}) and just a~fraction of them (experiments \exper{C} and \exper{D}). We have observed that with an increasing volume of identities the heterogeneous evaluation setup is on par with the homogeneous setup, that is, it does not matter what identities we learn the features on. One does not have to rely on the availability of all walkers for learning. This is particularly important for a system to aid video surveillance applications where encountered walkers never supply labeled data. Multiple occurrences of individual walkers can now be linked together even without knowing their actual identities.

\vspace{-10pt}%
\subsubsection*{Acknowledgments}
Authors thank to the anonymous reviewers for their detailed commentary and suggestions. Data used in this project was created with funding from NSF EIA-0196217 and was obtained from \url{http://mocap.cs.cmu.edu}. Our extracted database is available at \url{https://gait.fi.muni.cz/} to support results reproducibility.
\vspace{-22pt}%

\bibliographystyle{splncs03}
\bibliography{ref}
\end{document}